\documentclass{article}

\usepackage{microtype}
\usepackage{graphicx}
\usepackage{booktabs} 
\usepackage{amssymb}
\usepackage{xcolor}
\usepackage{colortbl}
\usepackage{graphicx}
\usepackage{multirow}
\usepackage{mathtools}
\usepackage{amsmath}
\usepackage{algorithm}
\usepackage{algorithmic}
\usepackage{amsfonts}
\usepackage{amsthm}
\usepackage{cite}
\usepackage{comment}

\usepackage{subfig}
\usepackage{caption}

\DeclareMathAlphabet{\mathcal}{OMS}{cmsy}{m}{n}

\usepackage{hyperref}

\usepackage[accepted]{icml2020}

\begin{document}

\twocolumn[
\icmltitle{%
Orthogonalized SGD and Nested Architectures for Anytime Neural Networks}



\icmlsetsymbol{equal}{*}

\begin{icmlauthorlist}
\icmlauthor{Chengcheng Wan}{uc}
\icmlauthor{Henry Hoffmann}{uc}
\icmlauthor{Shan Lu}{uc}
\icmlauthor{Michael Maire}{uc}
\end{icmlauthorlist}
\icmlaffiliation{uc}{University of Chicago, Chicago, IL, USA}
\icmlcorrespondingauthor{Chengcheng Wan}{cwan@uchicago.edu}


\vskip 0.3in
]



\printAffiliationsAndNotice{} 

\begin{abstract}
We propose a novel variant of SGD customized for training network architectures that support anytime behavior: such networks produce a series of increasingly accurate outputs over time. Efficient architectural designs for these networks focus on re-using internal state; subnetworks must produce representations relevant for both immediate prediction as well as refinement by subsequent network stages. We consider traditional branched networks as well as a new class of recursively nested networks. Our new optimizer, Orthogonalized SGD, dynamically re-balances task-specific gradients when training a multitask network. In the context of anytime architectures, this optimizer projects gradients from later outputs onto a parameter subspace that does not interfere with those from earlier outputs. Experiments demonstrate that training with Orthogonalized SGD significantly improves generalization accuracy of anytime networks.

\end{abstract}

\section{Introduction}
\label{sec:introduction}

\begin{figure}[t]
   \begin{minipage}[t]{\linewidth}
      \vspace{0pt}
      \centering
      \includegraphics[width=0.95\linewidth]{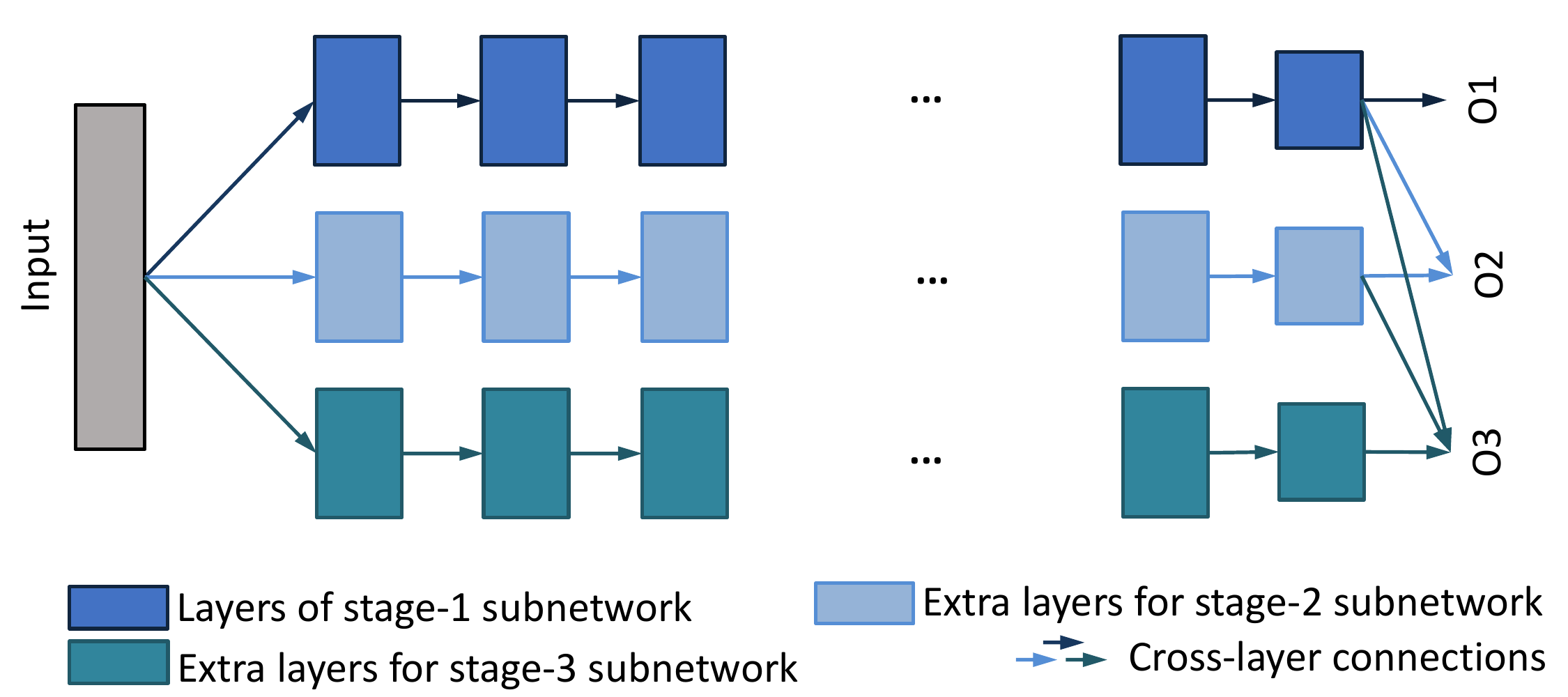}
      \vspace{-8pt}
      \caption{
         Ensemble of three different deep neural networks.
      }
      \label{fig:ensemble_example}
      \vspace{5pt}
   \end{minipage}
   \begin{minipage}[t]{\linewidth}
      \centering
      \includegraphics[width=0.95\linewidth]{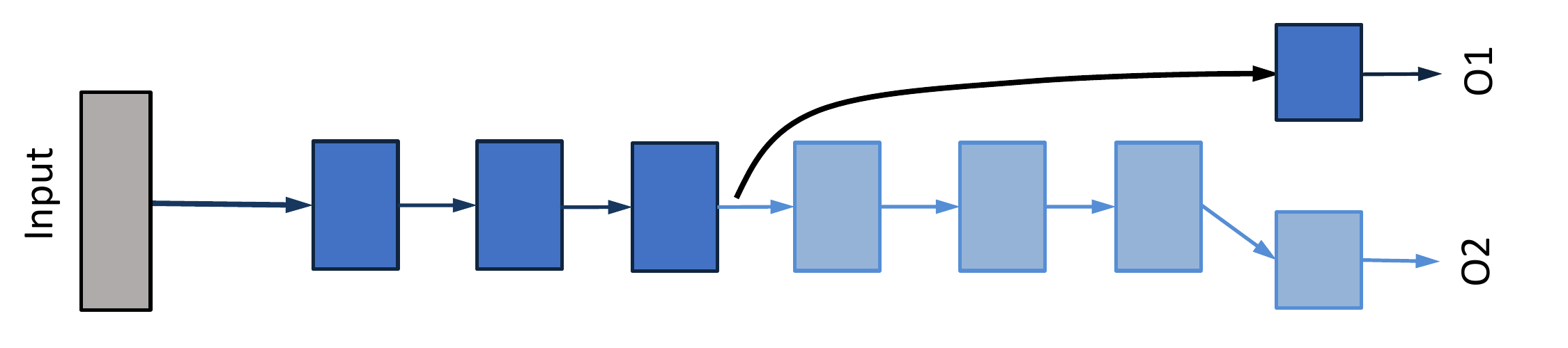}\\
      \vspace{-10pt}
      \caption{
         Cascade with branching outputs.  Networks are nested in depth,
         sharing a common trunk to which output branches attach.
         (\scriptsize{Box colors indicate in which inference stage a layer is
         introduced, as in Figure \ref{fig:ensemble_example}}).
      }
      \label{fig:depth}
      \vspace{5pt}
   \end{minipage}
   \begin{minipage}[t]{\linewidth}
      \centering
      \includegraphics[width=0.95\linewidth]{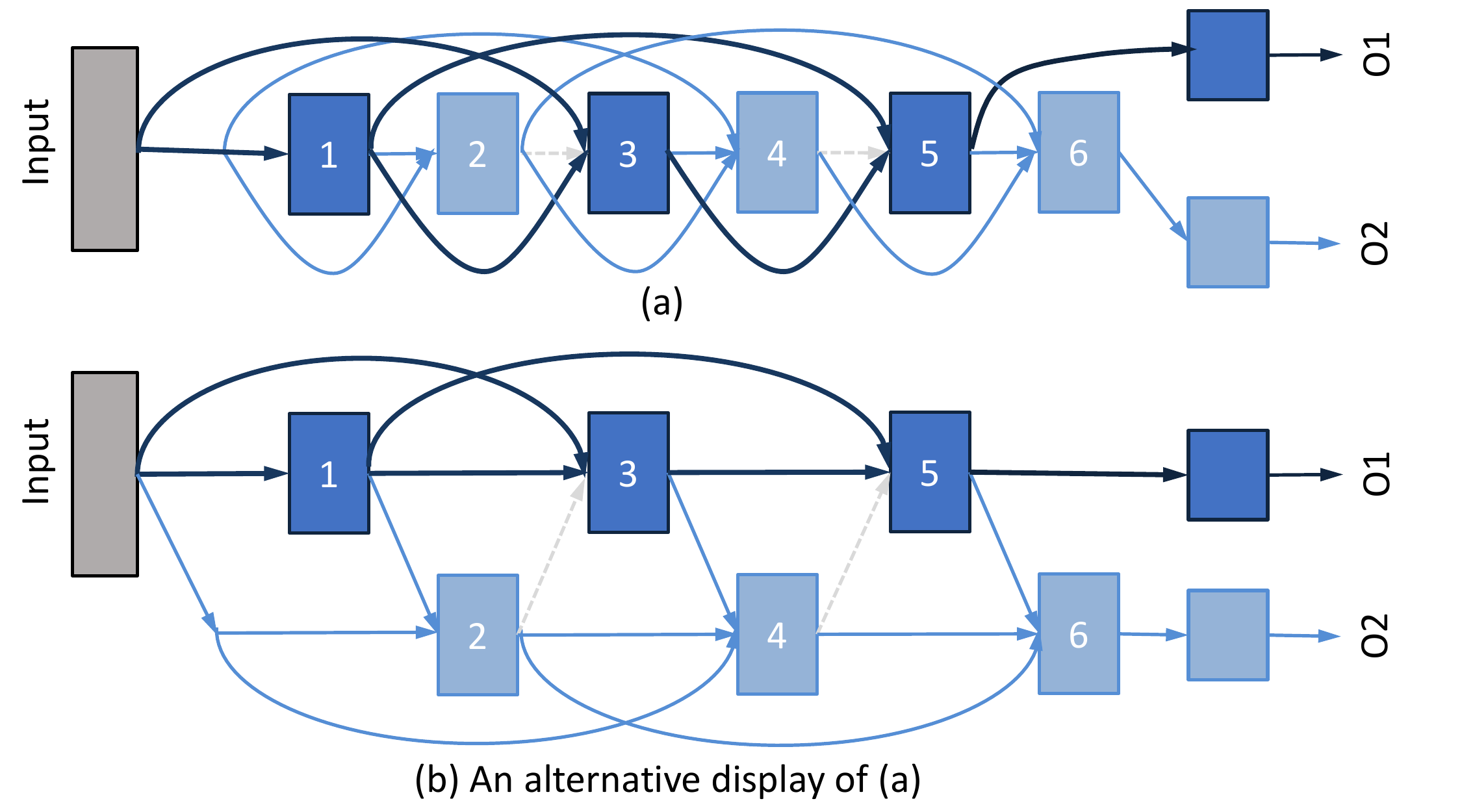}
      \vspace{-10pt}
      \caption{
         Our interlaced depth-wise nesting design for anytime networks.
         Recursive interlacing produces additional nested stages (beyond the
         two diagrammed here), each time doubling overall network depth.
         This design allows more synergistic coordination between networks
         than the branched topology shown in Figure~\ref{fig:depth}.
      }
      \label{fig:depth_sparse}
      \vspace{-12pt}
   \end{minipage}
\end{figure}

The accuracy of deep neural networks is affected by both their architecture and
overall size.  On one hand, improving architecture, via either principled
design~\cite{vgg,szegedy2015going,Resnet} or automated search
\cite{nas,enas,snas}, has been a major research focus; on the other, fixing a
particular architectural motif, increasing network size (\emph{i.e.}~width and
depth) provides a path to further improvement in accuracy---a trend prevalent
among dominant architectures in major application areas, including computer
vision~\cite{Resnet,zagoruyko2016wide} and natural language
processing~\cite{bert}.  The higher accuracy of larger networks comes at the
cost of increased compute requirements and longer inference latency.  Real
applications may be deployed on hardware (\emph{e.g.} mobile) and in use cases
(\emph{e.g.} interactive, real-time) that cap permissible network designs in
both compute and latency.

Network compression techniques~\cite{model_compression,han2015deep_compress}
can provide means of meeting some types of known, fixed resource constraints.
The pruning paradigm~\citep{han2015deep_compress} removes parameters
throughout a network, reducing memory and FLOP requirements, while distillation
approaches~\citep{han2015deep_compress,hinton2015distilling} can also shrink
network depth and thereby reduce latency.  However, none of these techniques
handle dynamic run-time environments, where applications may need to adjust
accuracy-compute-latency trade-offs in real-time.  For example, interactivity
requirements may present dynamically changing latency deadlines for predictions
\cite{dollar2011pedestrian,fowers2018configurable}; concurrently running
applications can compete for computation resources and power budget \cite{PCP}
in unpredictable ways \cite{zhuravlev2010contention}.  These environmental
factors could change while inference is executing \cite{lin2018architectural},
defeating attempts to statically schedule around them.

Anytime predictors \cite{zilberstein1996using} are a promising approach to
generating accurate inference results under dynamic latency and resource
constraints.  They produce complete outputs at multiple intermediate stages,
ameliorating penalties associated with failure to complete an inference
process.  If interrupted, an intermediate prediction---though less
accurate---substitutes for the originally desired output, thereby preventing
catastrophic failure.  General approaches to building anytime predictors
include ensembling~\cite{dietterich2000ensemble} multiple independent
predictors (Figure~\ref{fig:ensemble_example}) and reorganizing a standard
prediction pipeline into a cascade~\cite{zilberstein1996using}
(Figure~\ref{fig:depth}), both of which can be exploited to build anytime
variants of deep networks~\cite{wang2018anytime,lee2018anytime,huang2017multi,
mcgill2017deciding,branchynet}.

Unfortunately, anytime flexibility is not free.  Existing anytime network
design and training procedures sacrifice considerable accuracy and/or require
significant extra computation to produce intermediate predictions.  We reduce
these costs by introducing synergistic innovations across both anytime network
architectures and training procedures.

On the architectural aspect, we propose new structures for anytime neural
networks according to a principle of maximizing the potential for re-use of
intermediate state between successive stages.  A small network should not only
produce a quick output, but should also produce internal representations that
serve as valuable input to larger networks in subsequent stages.  We thus
design architectures so that connections between subnetworks in different
stages are aligned: they directly link corresponding pairs of layers across
stages, so as to allow subsequent subnetworks to refine previously computed
internal representations.

Maximizing the potential for cooperative refinement leads to a class of
recursively-nested anytime networks, in which later subnetworks fully contain
earlier ones, while growing in width, depth, or a combination thereof.
Compared to prior work~\citep{branchynet,huang2017multi} using branched cascade
architectures (Figure~\ref{fig:depth}), our alignment principle suggests a
depth-nesting approach based on interlacing subnetwork layers
(Figure~\ref{fig:depth_sparse}).  Conveniently, a recently proposed
improvement~\citep{sparsenet} to residual networks~\citep{Resnet}, though not
an anytime design itself, is amenable to adaptation into an interlaced,
depth-nested form; we defer details to Section~\ref{sec:architecture}.

\begin{figure}
   \centering
   \includegraphics[width=1.0\linewidth]{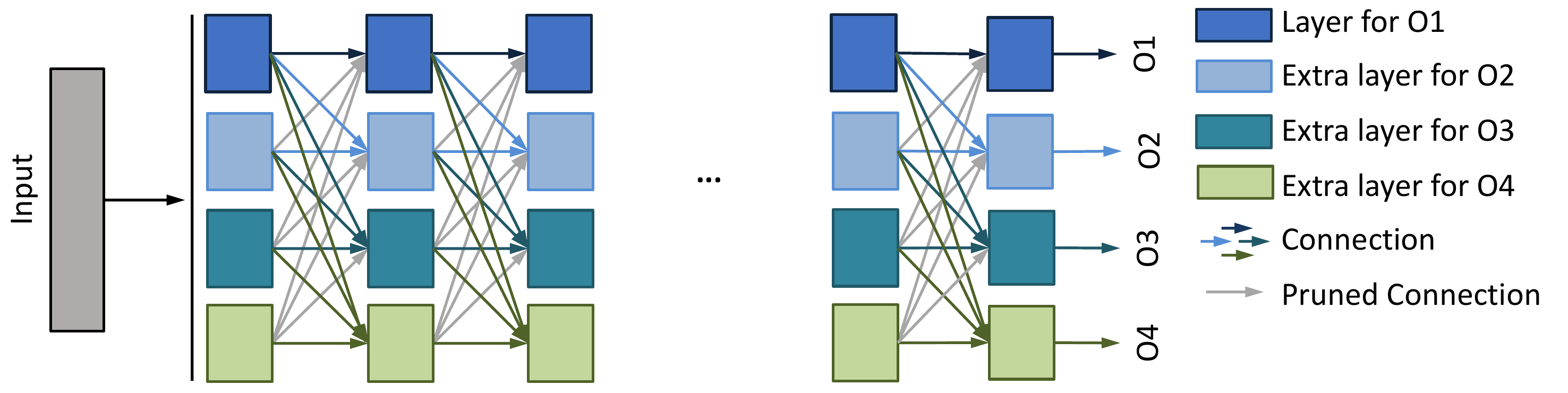}
   \vspace{-15pt}
   \caption{
      Width-wise nesting of deep networks.  Compared to a standard network,
      each layer is sliced into multiple layers (colored blocks, stacked
      vertically).  Each successive subnetwork includes another set of layer
      slices across the entire depth of the network.
   }
   \vspace{-5pt}
   \label{fig:width_base}
\end{figure}

Nesting networks width-wise is more straightforward
(Figure~\ref{fig:width_base}) and, unlike depth-nesting, previously-explored
solutions~\cite{lee2018anytime} agree with our alignment principle.  We do,
however, propose an improvement to width-nesting: subsequent stages should grow
exponentially, rather than linearly, in width.  This same scaling strategy
applies to our novel depth-nesting approach.  Experiments in
Section~\ref{sec:experiment} demonstrate the efficacy of interlaced
depth-nesting, as well as how exponential scaling of subnetwork size in
width- and/or depth-nested scenarios yields anytime networks that cover a
broad accuracy-compute trade-off curve.

Complementary to our architectural innovations, we propose a novel optimizer,
\emph{Orthogonalized SGD (OSGD)}, for training anytime neural networks.
Motivating OSGD is a view of anytime networks as a special-case of multitask
networks, combined with a desire to facilitate synergy between those tasks.
In addition to synergistic architectures, we want another type of synergy:
synergy in the optimization dynamics when training those multitask
architectures.

Here, the multiple tasks are different versions of the same task, restricted
to use only parameters within a particular subnetwork.  The partially-shared
structure of anytime networks, present in classical cascades and our nested
topologies, means that tasks may compete for use of shared parameters.  At the
same time, the highly-related nature of the tasks means that they may also act
as regularizers for one another.  OSGD provides a methodology for re-balancing
task interactions as they simultaneously pull on network parameters over the
course of training.  Section~\ref{sec:train} presents the technical details of
our re-balancing approach, which operates on a set of task-specific parameter
gradients.

While OSGD is general, with potential application to any multitask training
scenario, we restrict focus to anytime networks.  We observe dramatic
improvements in generalization accuracy when training anytime networks with
OSGD---a result that holds across the full spectrum of anytime network
architectures.  Training our fully-nested anytime networks with Orthogonalized
SGD sufficiently improves accuracy to the point of making such networks
competitive with standard designs lacking anytime flexibility.  Together, the
techniques we develop here provide a pathway toward endowing deep neural
networks with anytime flexibility at minimal overhead cost; as a consequence,
perhaps anytime designs should become the new default architectural schema.

\section{Related Work}
\label{sec:related}

\subsection{Dynamic Inference and Anytime Deep Networks}

Anytime algorithms have a well-established history, with computational
strategies applicable to many learning techniques~\cite{zilberstein1996using}.
Classic approaches include anytime decision trees using ensembles or
boosting~\cite{grubb2012speedboost,viola2004robust,xu2013cost,xu2012greedy}.
Past focus includes domains such as visual perception~\cite{viola2004robust},
which is now dominated by methods based on deep neural networks.  Adapting
anytime techniques to modern deep networks is of particular importance given
their computational demands and their potential use in real-time
safety-critical systems, such as autonomous vehicles~\cite{
huval2015empirical,Chen_2015_ICCV,Dosovitskiy17,teichmann2018multinet}.

Complementary to recent methods for statically reducing neural network model
size~\cite{han2015deep_compress,hinton2015distilling,SqueezeNet,
rastegari2016xnor,mobilenet}, another branch of investigation has focused on
reducing inference time in a dynamic, input-dependent manner~\cite{
figurnov2017spatially,mcgill2017deciding,andreas2017convolutional,
wu2018blockdrop,ren2018sbnet}.  These \textit{adaptive inference} methods skip
execution of parts of a network, based on an estimate of relevance computed
for each input; their goal is to minimize computation required for accurate
prediction on a per-example basis.  Here, the inference procedure changes
dynamically in response a network's input data.  However, these approaches do
not provide any mechanism for responding to environmental conditions that might
introduce transient resource constraints to the system.

Anytime methods, in contrast, do provide means of addressing such environmental
variability.  Specifically, they aim to introduce a degree of robustness to
dynamic environmental effects, at the possible cost of moderately increased
computation.  For example, a recent anytime design by \citet{wang2018anytime}
develops a prediction pipeline specifically for stereo depth estimation,
outputting images with increasing spatial resolution, an approach that may not
generalize to other domains.  Recent generic anytime approaches include several
\textit{cascade} designs~\cite{larsson2016fractalnet,branchynet,huang2017multi,
hu2019learning}, which grow subnetworks by depth, and a recent proposal
\cite{lee2018anytime} that grows by width, as discussed in
Section~\ref{sec:introduction}.  We compare our anytime designs with them
both conceptually and experimentally; Sections~\ref{sec:architecture}
and~\ref{sec:experiment} provide details.

\subsection{Multitask Training}

\citet{caruana1997multitask} suggests that multitask training can improve
generalization performance of machine learning systems.  Recent advances
highlight the ability of learned neural representations to transfer between
related tasks within particular domains, including vision
\cite{donahue2014decaf} and natural language processing \cite{bert}.
Fine-tuning previously trained networks for new tasks has become a standard
mode of operation.

Multitask learning has successes in multiple areas, including vision
\cite{kokkinos2017ubernet,eigen2015predicting,teichmann2018multinet}, natural
language processing \cite{collobert2008unified,hashimoto2016joint,
sogaard2016deep}, speech \cite{seltzer2013multi,wu2015deep}, and cross-domain
representations \cite{bilen2017universal}.  However, training such systems is
a non-trivial problem.  \citet{long2015learning,misra2016cross} use clustering
methods; \citet{yang2016deep} separate general and task-specific features;
\citet{kokkinos2017ubernet,normSGD} train all tasks with the same base network
and a few task-specific layers; \citet{kendall2018multi,hu2019learning} build
joint losses with adaptive weights.

Notably, \citet{kokkinos2017ubernet} finds it difficult to train a single
convolutional network for multimodal visual perception to the point of matching
the accuracy of more specialized task-specific networks.  This observation
runs contrary to the expectation that visual perception tasks should be highly
related to one another, sparking interest in developing methods to quantify
task interdependence \cite{zamir2018taskonomy}.

Similar in spirit to our approach, prior works have targeted changes
to optimizers to improve multitask network training.  This includes
NormSGD~\cite{normSGD}, which computes a parameter gradient per task, in
separate backpropagation passes.  These gradient vectors are then normalized
before summation, ensuring that each task exerts equal influence on network
parameters at every training iteration.  \citet{kendall2018multi} take another
approach to dynamically balancing task influence, allowing some slack in
relative task importance, provided it is justified by outsized gains in
accuracy across the task spectrum as a whole.  Our OSGD optimizer, like
NormSGD, attempts to dynamically re-balance task interactions.  However, OSGD
addresses a different kind of interaction, making it composable with most
existing optimizers; we also test an Orthogonalized NormSGD variant.

\subsection{Gradient Orthogonality}

Directly related to our method for reasoning about task interactions is the
work of \citet{farajtabar2019orthogonal}.  Like
\citet{farajtabar2019orthogonal}, we view task interactions through the
lens of their parameter gradient vectors.  That is, for a multitask network,
comparing per-task gradient vectors (computed via separate backpropagation
passes) tells us about task similarity: aligned gradients indicate synergy
between tasks, while opposing gradient directions suggest competition.
Orthogonal vectors suggest the tasks do not interfere with one another; they
each pull on a parameter subspace about which the other is agnostic.

\citet{farajtabar2019orthogonal} translate these intuitions into an algorithmic
approach for fine-turning that resists catastrophic forgetting.  Their goal is
to train an existing neural network to perform an additional task while
ensuring that it retains the ability to accurately perform a task it has
previously learned.  \citet{farajtabar2019orthogonal} model the parameter
subspace of the original task using a collection of parameter gradient vectors
for a sample of that task's training examples.  When training a new task, they
project its gradients onto the subspace orthogonal to that spanned by this
collection.  Even if using only a few hundred samples, this projection step
can be expensive.

As Section~\ref{sec:train} details, we instead orthogonalize gradient vectors
for the purpose of simultaneously training a multitask network from scratch.
We use orthogonalization to alter optimizer dynamics in an online, approximate
manner.  Unlike \cite{farajtabar2019orthogonal}, we do not have access to a
gradient collection modeling the true parameter subspace of a trained task.
Surprisingly, using just a single gradient vector per (partially trained) task,
computed over a minibatch, proves to be sufficient.  Our OSGD algorithm
dynamically re-balances task interactions online, and at far lower
computational expense.

\section{Anytime Network Architectures}
\label{sec:architecture}

\subsection{Baselines}
\label{sec:design_baselines}

\textbf{Cascade networks} add early exit branches from the main network
pipeline \cite{huang2017multi,mcgill2017deciding,branchynet,hu2019learning}.
As illustrated in Figure~\ref{fig:depth}, early outputs are generated
without traversing later pipeline stages---which tend to capture high-level
input features---leading to large accuracy loss for early outputs.  Cascading
also requires extra computation on every early output path to convert the
intermediate representation of that layer to a suitable output.  Training such
cascades puts conflicting pressure on layers that serve heterogeneous branches
(\emph{e.g.}, a block can be connected to both an output layer and another
intermediate layer in Figure~\ref{fig:depth}).

\textbf{Equal-width nested networks} split a neural network into $n$
equal-width horizontal stripes \cite{lee2018anytime}, as
Figure~\ref{fig:width_base} illustrates.  Each stripe executes sequentially.
Compared to branched cascades, this configuration offers more intermediate
state reuse opportunities across subnetworks.  Compared to a regular network of
similar size, some connections are removed, as one cannot have edges from
latter stripes to earlier stripes (gray edges in Figure \ref{fig:width_base}).
Furthermore, although increasing network width increases accuracy, benefits do
not typically scale linearly with network size.  Consequently, the design in
Figure~\ref{fig:width_base} may produce intermediate results with suboptimal
accuracy-latency trade-offs.

\subsection{Design Principles}
\label{sec:design_prin}

Three observations guide our anytime architecture designs.

\textbf{Grow \emph{both} width and depth.}
Accuracy improves with both deeper (more layers)~\cite{vgg,szegedy2015going,
Resnet} and wider (more neurons per layer)~\cite{zagoruyko2016wide,bert}
designs.  Moreover, this trend is especially pronounced within an architecture
family: if residual networks (ResNets) are preferred for a particular domain,
a 50-layer ResNet will deliver better accuracy than a 34-layer
ResNet~\cite{Resnet}.  Consequently, we develop freely composable recipes
for nesting networks in width and depth.

\textbf{Grow fast.}
Although accuracy typically improves with network size, this improvement
usually falls off as size increases; logarithmic scaling of improvements are
a common result.  Consequently, we increase network size exponentially from one
stage to the next.  This places output predictions at useful discrete accuracy
steps along a trade-off curve.  This design choice also minimizes cut
connections when transforming a standard network into an anytime version.

\textbf{Reuse intermediate state.}
We improve efficiency by fully reusing {\it internal} activation states of
earlier subnetworks to bootstrap later subnetworks.  Recent investigations of
learned representations \cite{unrolled,resnet_iter} suggest that layers within
very deep networks might be learning to perform incremental updates, slowly
refining a stored representation.  Deeper networks might refine their
representations more gradually than shallower networks.  By aligning layers of
different subnetworks trained for the same task, according to the relative
depth in their own subnetwork, we might jump-start computation in larger
subnetworks.

\subsection{Nested Anytime Network Architectures}
\label{sec:design_any}

Our design consists of a sequence of {\it fully nested} subnetworks: the first,
$D_1$, is completely contained within the second, $D_2$, which is a subpart of
$D_3$, \emph{etc.}  Going from $D_{i}$ to $D_{i+1}$, our scheme permits growing
the network in width, depth, or both.  Our anytime networks also have the
following properties:
(1) \emph{pipeline structure}: Every subnetwork $D_i$ follows the usual
pipeline structure of a traditional neural network (as opposed to the
branching present in cascade networks);
(2) \emph{aligned feed forward}: Outputs of internal layers of a smaller
subnetwork are forwarded to deeper layers of the same subnetwork, as well as
internal layers of the larger network most appropriate for consuming their
signals, maximizing data reuse (\emph{i.e.}, connections are purely
feed-forward in depth or nesting level);
(3) \emph{exponential size scaling}: The sizes of subnetworks increases
exponentially so later outputs offer meaningful accuracy improvements over
earlier ones.

The difference between an independent non-nested network and every subnetwork
$D_i$ is that some neurons in an outer nesting level will not feed backward
into neurons in inner nesting levels.  Dropping these connections slightly
reduces the compute load and parameter count, slightly shifting the network's
position on the latency-accuracy curve, which is inconsequential in the
big picture of an anytime network that populates the trade-off curve with many
nested subnetworks.

\subsubsection{Depth Nesting}
\label{sec:architecture_d}

We \emph{interlace} layers following the same pipeline structure as the
original network.  As illustrated in Figure~\ref{fig:depth_sparse}, we
partition a traditional network into odd and even layers.  We create a
shallower subnetwork consisting of only the odd numbered layers to produce the
first intermediate result, and nest it within the full network, which has
double the depth.  Recursively applying this process, we create a sequence of
interlaced networks that repeatedly double in depth.

This depth-nesting strategy applies only to networks satisfying an additional
architectural requirement.  Notice, in Figure~\ref{fig:depth_sparse}, the
presence of additional skip connections between layers, even in the basic,
non-nested network.  Indeed, within any network in the sequence, we must have
that each layer connects directly to any other layer separated in depth by a
power of 2.  Fortunately, this power-of-2 skip-connection design is exactly the
SparseNet architecture~\cite{sparsenet}, which is a state-of-the-art variant of
ResNet~\cite{Resnet} (or DenseNet~\cite{densenet}) convolutional networks.

\subsubsection{Width Nesting}
\label{sec:architecture_w}

\begin{figure}
   \centering
   \includegraphics[width=1.0\linewidth]{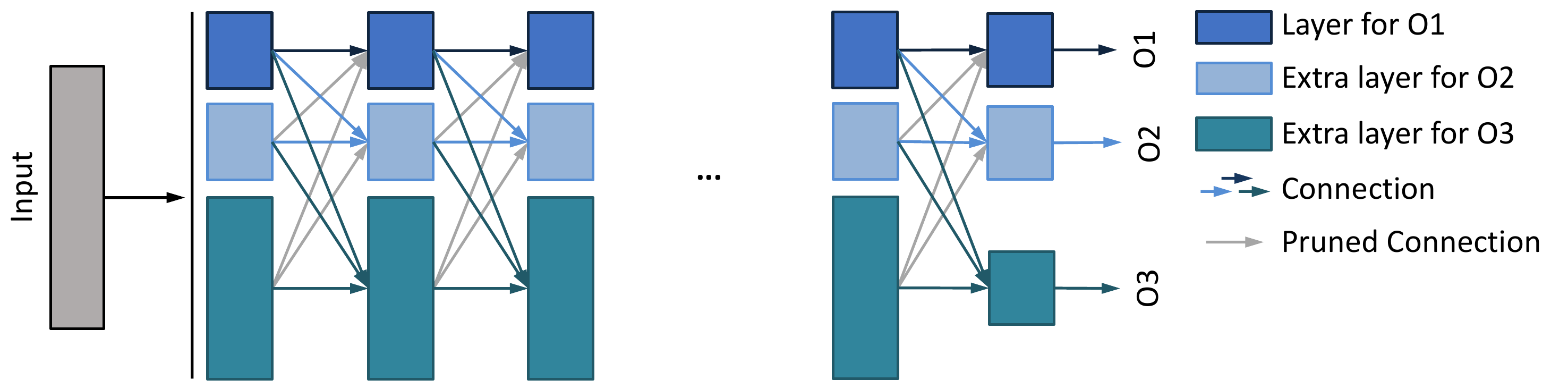}
   \vspace{-0.15in}
   \caption{Our width-wise nesting of subnetworks.}
   \label{fig:width}
\end{figure}

Similar to a \citet{lee2018anytime}, our width-nesting strategy divides a
network into $M$ horizontal stripes, with the $i$-th subnetwork including all
the neurons inside the first $i$ stripes.  Different from this prior work, we
use a power-of-2 sequence for stripe widths, as Figure~\ref{fig:width} depicts.

If the first subnetwork $D_1$ contains $w$ neurons in one layer, $D_i$ contains
$w \times 2^{i-1}$ neurons in the corresponding layer.  This choice creates a
good trade-off curve for accuracy and latency.  Additionally, this exponential
stripe-width split causes far fewer edges to be pruned than the even-width
split in prior work~\cite{lee2018anytime}.  All the connections from a
later-stripe neuron to an earlier-stripe neuron need to be pruned, like those
upward gray edges in Figures~\ref{fig:width_base} and~\ref{fig:width}.

\subsubsection{Combining Depth and Width Nesting}
\label{sec:architecture_dw}

\begin{figure}
   \centering
   \includegraphics[width=1.0\linewidth]{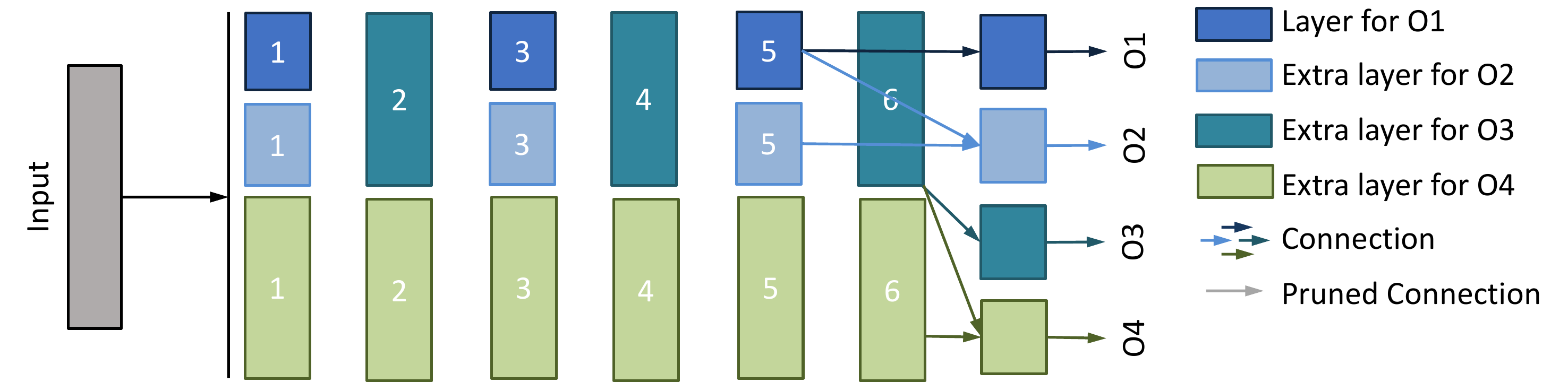}
   \vspace{-0.3in}
   \caption{%
      Our width-depth nesting that alternates growing width and depth.
      Connections across intermediate layers are hidden.%
   }%
   \label{fig:wd_example}
\end{figure}

\begin{figure}
   \centering
   \includegraphics[width=1.0\linewidth]{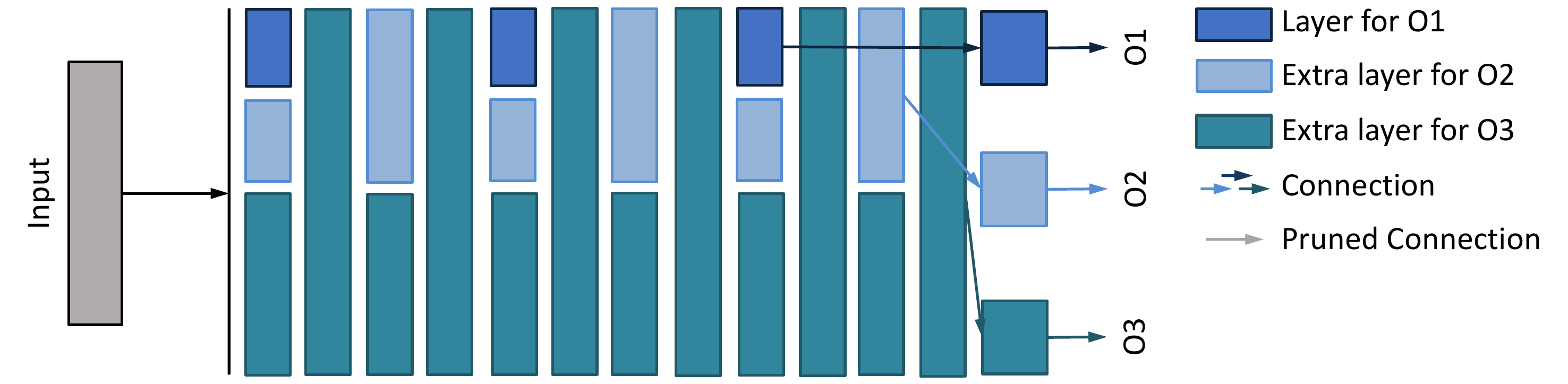}
   \vspace{-0.3in}
   \caption{%
      Our width-depth nesting that grows width and depth
      simultaneously. Connections across intermediate layers are hidden.%
   }%
   \label{fig:wd_example2}
\end{figure}

Our width and depth nesting designs can be easily combined in arbitrary order:
depth then width, width then depth, or combinations thereof.  When growing
depth, interlaced layers are added.  When growing width, all layers double
their filter count.  Figure~\ref{fig:wd_example} illustrates growth by
alternating width and depth: subnetwork-1 (dark blue layers) grows to
subnetwork-2 by extending its width (light blue layers), then grows to
subnetwork-3 by extending depth (green layers), and then to subnetwork-4 by
extending width again (light green layers).  Figure~\ref{fig:wd_example2}
illustrates an alternative of simultaneous growth in width and depth.

\section{Optimization Strategies}
\label{sec:train}

Every anytime network (using our architecture or others) faces a multitask
training challenge: simultaneous optimization of losses attached to outputs
of multiple subnetworks.  In this section, we propose Orthogonalized SGD
(OSGD), a new optimizer for training multitask deep networks, which, as
Section~\ref{sec:experiment} demonstrates, proves particularly effective when
applied to anytime networks.

Prior to presenting Orthogonalized SGD, we describe several optimization
strategies appropriate for anytime networks.  Performance of these baselines
provides a benchmark against which to compare OSGD.  The optimizers we consider
vary in how they address two crucial issues:
(1) how to weight losses from multiple outputs, and
(2) how to share network parameters between multiple tasks.

Our simplest baseline optimizer, a greedy training approach, couples loss
weighting and parameter sharing.  By training an anytime network in a greedy
stage-wise fashion, it only ever has one loss to consider and one active set of
parameters.  Specifically, the greedy optimizer trains the smallest subnetwork
to produce the first output.  It then freezes parameters of that subnetwork and
trains the next subnetwork, repeating in such fashion until the last output is
trained.  The anytime network's parameters are thus partitioned into subsets
that are optimized exclusively for each output.  However, greedy training
potentially harms later outputs as they cannot influence parameters in the
portion of the network they share with earlier outputs.

Decoupling loss weighting and parameter sharing strategies, we span a larger
optimizer design space.  For loss weighting, we consider a fixed linear
combination, as well as a dynamic normalization scheme which equalizes the
influence of each task at every training step.  SGD and NormSGD are the
corresponding optimizers we consider; they represent the default parameter
sharing strategy.  OSGD and its normalized variant manage parameter sharing
in a more sophisticated manner.  They give some tasks preferential pull on
certain parameter subsets, while still allowing other tasks some influence.
This prioritization process depends on the gradients from each task,
thereby inducing entirely different learning dynamics.

\subsection{Definitions and Preliminaries}

Training a nested anytime network is an instance of multitask learning, where
the tasks are solving the same problem with different network components
(specifically, the outputs of different subnetworks).

Let $w_1 \in R^{d_1}, w_2 \in R^{d_2}, \cdots, w_n \in R^{d_n}$ be the weights
of the nested networks, where $d_1 < d_2 < \cdots < d_n$ and
$w_1 \subsetneq w_2 \subsetneq \cdots \subsetneq w_n$.  We define other
symbols as follows:
\vspace{-20pt}
\begin{itemize}
   \setlength\itemsep{0em}
   \item $W$: weight for the whole network, equivalent to $w_n$
   \item $L_i$: the loss of subnetwork $D_i$ ($D_i$ has weights $w_i$)
   \item $g_i$: the gradient of weights $w_i$ from loss $L_i$.
   \item $g_i^j$: the gradient of weights $w_j \setminus w_{j-1}$ from loss
               $L_i$, where $j \leq i$; Note that $g_i^j$ is a subset of $g_i$.
   \item $k_i$: the importance of loss $L_i$
   \item $C$: a constant value for normalization
\end{itemize}

\subsection{Baseline Optimizers}

\paragraph{Greedy (stage-wise SGD)}
The greedy optimizer separates weights (parameters) into stages corresponding
to the structural organization of the anytime network:
$w_1, (w_2{\setminus}w_1), (w_3{\setminus}w_2), \ldots, (w_n{\setminus}w_{n-1})$.
It trains the first stage to achieve the lowest possible loss for $L_1$.
It then freezes all weights inside the first stage, and trains the second
subnetwork, which contains the first two stages.  This then continues for the
third, and so on.  (Algorithm~\ref{alg:greedy})

This strategy typically yields high accuracy for the smallest network, as
parameters $w_1$ are entirely dedicated to that network's output task.
However, it pays for this guarantee by having suboptimal coordination with
later output tasks, as they cannot adjust $w_1$ for their own benefit.

\begin{algorithm}[tp]
\caption{Greedy stage-wise multitask training}
\label{alg:greedy}
\begin{algorithmic}[1]
   \STATE $w_0 \Leftarrow \emptyset$
   \FOR {$i=1$ \textbf{to} $n$}
      \STATE Initialize weights $w_i \setminus w_{i-1}$
      \FOR {$t=0$ \textbf{to} \emph{max\_train\_steps}}
         \STATE Compute $L_i(t)$ \hfill [forward pass]
         \STATE $g_i(t) \Leftarrow \nabla_{w_i \setminus w_{i-1}} L_i(t)$
         \STATE Update $W(t) \mapsto W(t+1)$ using $g_i(t)$
      \ENDFOR
   \ENDFOR
\end{algorithmic}
\end{algorithm}

\paragraph{SGD}
With standard stochastic gradient descent (SGD), each task contributes a term
to the overall loss in the forward path.  A straight-forward approach is to
compute a weighted average of these terms, and then conduct standard backprop:
\begin{equation}
   L = \frac{\sum_{i=1}^{n} k_i L_i}{ \sum_{i=1}^{n}k_i }.
\end{equation}
The weighted average function could be chosen to match known operating
environment characteristics (\emph{e.g.}, the time budget and latency on a
target machine), which provides great flexibility.

\paragraph{NormSGD}
Globally normalizing gradients can help to dynamically balance the importance
of different tasks \cite{normSGD,chen2017gradnorm}.  Doing so directly modifies
gradient magnitudes to balance the contribution of each task at each training
iteration.  This effect is not equivalent to any static reweighting of losses.
For gradient $g_i \in R^{d_i}$, it is normalized as:
\begin{equation}
   g_i = \frac{g_i}{\left\lVert g_i(t) \right\rVert} \cdot \sqrt{d_i} \cdot C
\end{equation}
where $\sqrt{d_i}$ compensates for the fact that we are working with
subnetworks of different size.  This multiplicative term gives each subnetwork
equal influence upon the parameters it shares with other subnetworks.  When
updating parameters, we use an average of the (normalized) gradients of the
subnetworks in which they participate.  To allow use of NormSGD with
standard learning rate schedules, we multiply the gradient by a constant
value $C$, which we set using calibration experiments.

\begin{algorithm}[!htp]
\caption{Orthogonalized SGD: A multitask variant of SGD with optional
dynamic \emph{normalization} of task influence.}
\label{alg:unified}
\begin{algorithmic}[1]
   \STATE Initialize weights $W$
   \FOR {$t=0$ \textbf{to} \emph{max\_train\_steps}}

      \STATE Compute $L_i(t)~\forall i,~s.t.~1 \leq i \leq n$ \hfill [forward pass]
      \STATE $g(t) \Leftarrow \textbf{0}$

      \FOR {$i=1$ \textbf{to} $n$}
         \STATE $g_i(t) \Leftarrow$ $\nabla_{w_i} L_i(t)$
         \IF{\emph{normalizing}}
            \STATE $g_i(t) \Leftarrow g_i(t) / \left\lVert g_i(t) \right\rVert \cdot \sqrt{d_i} \cdot C$
         \ENDIF
      \ENDFOR

      \FOR {$i=1$ \textbf{to} $n$}
         \STATE $h_i(t) \Leftarrow \sum_{j=1}^{i-1} \textit{proj}_{g_j(t)}g_i(t)$
         \STATE $g_i(t) \Leftarrow g_i(t) - h_i(t)$
         \STATE $g(t) \Leftarrow g(t) + g_i(t)$
      \ENDFOR

      \STATE Update $W(t) \mapsto W(t+1)$ using $g(t)$

   \ENDFOR
\end{algorithmic}
\end{algorithm}

\subsection{Orthogonalized SGD (OSGD)}

Our novel optimizer, Orthogonalized SGD, dynamically re-balances task-specific
gradients in a manner that prioritizes the influence of some losses over
others.  Given loss-specific gradient vectors $g_1, g_2, \ldots, g_n$,
Orthogonalized SGD projects gradients from later outputs onto the parameter
subspace that is orthogonal to that spanned by the gradients of earlier
outputs.  As a result, subsequent outputs do not interfere with how earlier
outputs desire to move parameters.  For example, the retained component of the
gradient of $w_2$ is
\begin{equation}
   g'_2 = g_2 - \textit{proj}_{g_1}{g_2},
\label{eq:prioritize}
\end{equation}
where $\textit{proj}_{A}{B}$ refers to projecting vector $B$ onto $A$.
$g'_2$ is orthogonal to $g_1$, and thus updating $w_1$ in the direction of
$g'_2$ minimizes interference with the optimization of loss $L_1$.

Algorithm~\ref{alg:unified} provides a complete presentation of both
Orthognolized SGD and an orthogonalized variant of NormSGD.  Note that for
anytime networks, per-task gradient vectors are padded with zero entries for
any parameters not contained in the corresponding subnetwork.  For example,
$g_1$ pads zeros to $w_2 \setminus w_1$, so the part of $g_2$ specific to the
second subnetwork will be unaffected by Equation~\ref{eq:prioritize}.

More generally, OSGD can be used with any priority ordering of tasks; the
priority order need not correspond to the order in which outputs are generated
by an anytime network.  Algorithm~\ref{alg:unified} is valid for any shuffling
of losses, regardless of the underlying network architecture.  Choosing a
priority order determines the sequencing of gradient projection steps, thereby
changing which tasks are given preferential influence over network parameters.

As part of the following section, we discuss why an early-to-late priority
order delivers excellent results for anytime networks.  We also explore how
different task priority ordering can serve as a tool for customizing network
behavior towards specific time-dependent prediction utility curves.

\section{Experiments}
\label{sec:experiment}

\subsection{Methodology}
\label{sec:meth}

We begin with evaluation using the CIFAR-10 dataset \cite{cifar}.  All networks
are trained for 200 epochs, with learning rate decreasing from 0.1 to 0.0008.
We train every network 3 times, and report the average and standard deviation
of its validation error.

We evaluate all five optimization strategies from Section \ref{sec:train}:
Greedy stage-wise training, SGD, OSGD, and the normalized variants of both
SGD and OSGD.  We set $C={1}/{2}$ and use a constant loss importance for SGD
and NormSGD, as these settings provide the best results.

We evaluate six different anytime network architectures: four novel designs of
our own and two prior designs.  Our designs include:
(1) depth-nesting applied to Sparse ResNet-98 \cite{sparsenet}
(Figure \ref{fig:depth_sparse}),
(2) width-nesting applied to ResNet-42 \cite{Resnet} (Figure \ref{fig:width}),
(3) alternating width-depth nesting (Figure \ref{fig:wd_example}), and
(4) simultaneous width-depth nesting (Figure \ref{fig:wd_example2}), with
the latter two applied to Sparse ResNet-98 \cite{sparsenet}.

The two previous designs represent the state-of-the-art depth-growing anytime
design, referred to as {\it EANN} (Figure \ref{fig:depth}) and width-growing
anytime design, referred to as {\it Even-width} (Figure \ref{fig:width_base}).
In {\it EANN}, we apply the cascade-based approach \cite{hu2019learning} to
Sparse ResNet-98, which grows depth exponentially and assembles an output
branch every $k \cdot 2^i (i=1,2,...)$ layers.
In {\it Even-width}, we apply the idea of recently proposed even-sized
width-nested architecture \cite{lee2018anytime} to ResNet-42.

\subsection{Evaluation of Optimization Strategies}
\label{sec:exp_train}

\definecolor{Gray}{gray}{0.85}
\newcolumntype{g}{>{\columncolor{Gray}}c}

\begin{table}[t]
\centering
\small
\setlength{\tabcolsep}{2.5pt}
\begin{tabular}{@{}c|ggcgc@{}}
\hline
Stage$_\text{size}$&Greedy&SGD& OSGD& SGD$_\text{Norm}$&OSGD$_\text{Norm}$\\
\hline
\multicolumn{6}{c}{Our Depth Nested Sparse ResNet-98}\\
\hline
1$_{d1}$                                                         & {9.6}~(0.2)    & 9.8 (0.1) & 10.0~(0.3) & 10.0~(0.2)                                           & 10.7~(0.2)                                              \\
2$_{d2}$                                                         & 9.3~(0.3)    & {8.3}~(0.3) & 8.4~(0.1)  & 8.6~(0.4)                                            & 8.5~(0.3)                                               \\
3$_{d4}$                                                        & 9.2~(0.3)    & 7.7~(0.3) & {7.4} (0.1)  & 8.1~(0.3)                                            & 7.6~(0.1)                                               \\
4$_{d8}$                                                        & 9.1~(0.2)    & 7.2~(0.4) & {6.6}~(0.1)  & 8.0~(0.2)                                            & 6.9~(0.1)                                               \\
\hline
\multicolumn{6}{c}{Our Width Nested ResNet-42}\\
\hline
1$_{w1}$                                                         & {10.2}~(0.1)   & 12.2 (0.2) & 12.3 (0.1) & 12.3 (0.3)                                           & 12.7 (0.1)                                              \\
2$_{w2}$                                                         & 9.9~(0.2)    & 10.1 (0.1) & {8.9} (0.2)  & 10.1 (0.2)                                           & 9.6 (0.4)                                               \\
&-&-&-&-&-\\
3$_{w4}$                                                         & 9.2~(0.2)    & 9.8 (0.3)  & {7.3} (0.3)  & 10.1 (0.2)                                           & 7.4 (0.2)                                               \\
\hline
\multicolumn{6}{c}{Our (Alternating) Width-Depth Nested Sparse ResNet-98}\\
\hline
1$_{w1d1}$                                                         & {18.5}~(0.1)   & 31.4 (0.6) & 28.3 (0.4) & 30.7 (0.4)                                           & 28.1~(0.5)                                              \\
2$_{w2d1}$                                                          & 16.5~(0.1)   & 15.6~(0.2) & 14.8 (0.2) & 15.5~(0.3)                                           & {14.7}~(0.4)                                              \\
3$_{w2d2}$                                                          & 15.9~(0.2)   & 15.5~(0.2) & {13.4}~(0.3) & 15.4~(0.2)                                           & 14.1~(0.2)                                              \\
4$_{w4d2}$                                                          & 15.7~(0.4)   & 10.4~(0.4) & {8.6}~(0.3)  & 10.4~(0.2)                                           & 9.4~(0.2)                                               \\
5$_{w4d4}$                                                          & 15.6~(0.3)   & 8.8~(0.3)  & {6.8}~(0.2)  & 8.9~(0.3)                                            & 7.4~(0.2)                                               \\
\hline
\multicolumn{6}{c}{Our (Simultaneous) Width-Depth Nested Sparse ResNet-98}\\
\hline
1$_{w1d1}$                                                         & {18.5}~(0.1)   & 28.0 (0.2) & 26.2 (0.1) & 29.1 (0.5)                                           & 26.7~(0.5)                                              \\
2$_{w2d2}$                                                         & {11.4}~(0.1)   & 15.0~(0.3) & 13.1~(0.1) & 15.6~(0.5)                                           & 14.5~(0.4)                                              \\
3$_{w4d4}$                                                         & 8.6~(0.4)   & 8.5~(0.3)  & {6.8}~(0.3)  & 9.0~(0.2)                                            & 7.4~(0.1)                                               \\
\hline
\end{tabular}
\vspace{-0.1in}
\caption{%
   CIFAR-10 error rates, the lower the better, of our anytime networks with
   different optimization strategies.  Numbers in parentheses are standard
   deviations.  Size subscripts indicate the subnetwork width or depth
   normalized to that of the first-stage subnetwork.  OSGD consistently
   improves over SGD and, compared to both SGD and Greedy stage-wise training,
   achieves dramatically lower error for later outputs.%
}%
\vspace{-0.1in}
\label{tab:training_our}
\end{table}

Tables~\ref{tab:training_our} and~\ref{tab:training_base} show the validation
error rates of applying five different optimizers to different anytime
networks.  Overall, our Orthogonalized SGD and its normalized variant perform
the best, capable of achieving high accuracy for later outputs of an anytime
network without significantly reducing the accuracy for earlier outputs.

Compared with SGD, OSGD consistently achieves higher accuracy for the last two
subnetworks across {\it all} six anytime designs, while maintaining similar or
better accuracy for early subnetworks.  Switching from SGD to OSGD drops the
last-stage error rates from 7.2, 9.8, 8.8 and 8.5 down to 6.6, 7.3, 6.8 and
6.8 across the four anytime networks in Table~\ref{tab:training_our}.
While the greedy training strategy offers the highest accuracy for the first
intermediate result of all anytime networks, it falls far behind OSGD for
later-stage results.

The improvement offered by OSGD is striking, yet somewhat counterintuitive.
These experiments give earlier outputs high priority than later outputs.
OSGD is prioritizing the influence that gradients of smaller subnetworks have
on the training dynamics, but it is the outputs of larger subnetworks that
most improve in accuracy.

A possible explanation for this curious behavior stems from the fact that the
multiple tasks in anytime networks are highly related.  In particular, in
a well-architected anytime network, different output tasks might exert a
beneficial regularization effect on one another.  OSGD, by prioritizing task X
over task Y in such a network then triggers two effects:
\begin{itemize}
   \vspace{-5pt}
   \item{It allocates parameters to task X instead of task Y.}
   \vspace{-5pt}
   \item{It decreases the regularization influence of task Y on task X, while
   simultaneously increasing the regularization influence of task X on task Y.}
\end{itemize}
Individually, these effects move the relative accuracy of task X and Y in
opposite directions.  As they are coupled, we observe only the net result.
Regularization interaction being the stronger effect would explain the
behavior of anytime networks trained with OSGD.  But, further investigation is
required before confidently adopting this explanation.

\begin{figure}
   \centering
   \vspace{10pt}
   \includegraphics[width=0.95\linewidth]{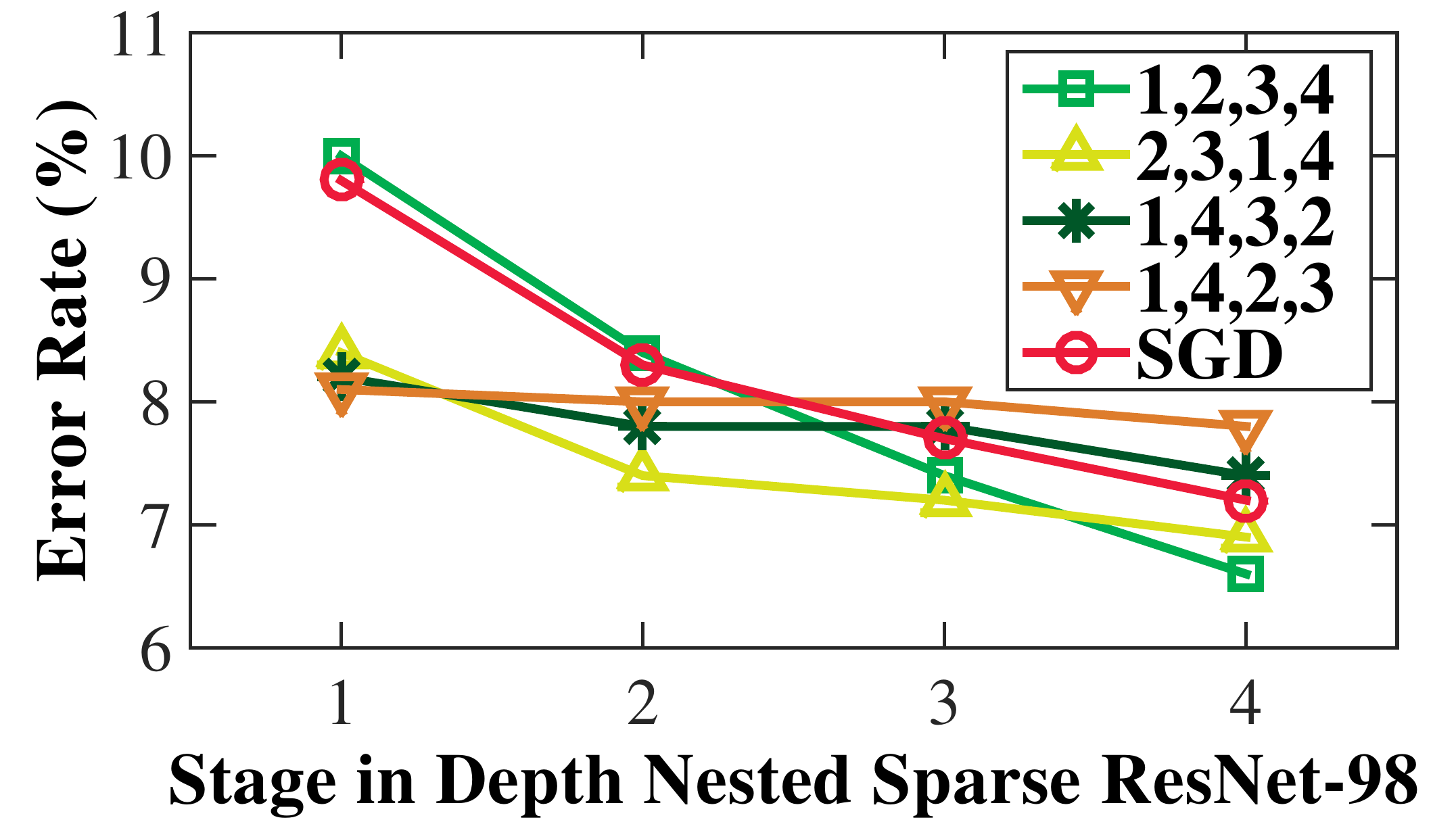}
   \vspace{-5pt}
   \caption{%
      Re-prioritizing outputs within the same anytime architecture, OSGD
      training focuses on reducing error at certain stages at the expense of
      others.  Shown are results for OSGD training with four different stage
      priority orders, as well as an SGD baseline.%
   }%
   \label{fig:priority}
\end{figure}

Figure~\ref{fig:priority} shows that, by changing the task priority order, we
can use OSGD to improve the accuracy of intermediate-stage anytime outputs at
the expense of later-stage outputs.  Compared to the standard priority order
(green line), giving intermediate outputs higher priority (yellow line) reduces
their error while slightly increasing error for the final output.  This
capability could be used to target an anytime system around a critical
prediction time window.

\begin{table}[t]
\centering
\small
\setlength{\tabcolsep}{2.0pt}
\begin{tabular}{@{}c|ggcgc@{}}
\hline
Stage$_\text{size}$&Greedy&SGD& OSGD& SGD$_\text{Norm}$&OSGD$_\text{Norm}$\\
\hline
\multicolumn{6}{c}{EANN Cascade Sparse ResNet-98}\\
\hline
1$_{d1}$                                                         & {9.3}~(0.1)    & 11.7 (0.3) & 11.6 (0.3) & 12.4 (0.1)                                           & 12.1 (0.4)                                              \\
2$_{d2}$                                                         & {9.2}~(0.3)    & 11.1 (0.1) & 10.9 (0.2) & 12.0~(0.1)                                           & 11.2 (0.1)                                              \\
3$_{d4}$                                                        & 8.8~(0.3)    & 8.5 (0.2)  & {8.0} (0.1)  & 9.2 (0.2)                                            & 9.0 (0.2)                                               \\
4$_{d8}$                                                        & 8.5~(0.3)    & 6.5 (0.2)  & {6.4}~(0.2)  & 8.0 (0.1)                                            & 7.6 (0.1)                                               \\
\hline
\multicolumn{6}{c}{Even-Width Nested ResNet-42}\\
\hline
1$_{w1}$                                                         & {10.2}~(0.04)   & 12.7 (0.2)  & 13.9 (0.1) & 12.6 (0.1)                                           & 13.5 (0.2)                                              \\
2$_{w2}$                                                         & {9.9}~(0.3)    & 10.2 (0.3)) & 10.7 (0.2) & 10.6 (0.1                                            & 10.8 (0.1)                                              \\
3$_{w3}$                                                         & 9.9~(0.4)    & 10.0 (0.1)  & {8.3} (0.02) & 10.5 (0.1)                                           & {8.3} (0.01)                                              \\
4$_{w4}$                                                         & 9.8~(0.2)    & 9.9 (0.1)   & {8.3} (0.1)  & 10.4 (0.1)                                           & {8.3} (0.1)                                               \\
\hline
\end{tabular}
\vspace{-0.1in}
\caption{
   CIFAR-10 error rates of previous anytime networks with different
   optimization strategies.  As in Table~\ref{tab:training_our}, OSGD
   offers benefits compared to other optimizers.
}
\vspace{-0.2in}
\label{tab:training_base}
\end{table}

\subsection{Evaluation of Nested Architectures}
\label{sec:exp_dnn}

\begin{figure}
   \begin{minipage}{1.0\linewidth}
      \centering
      \begin{minipage}[t]{0.32\linewidth}
         \vspace{0pt}
         \centering
         \includegraphics[width=1.0\linewidth]{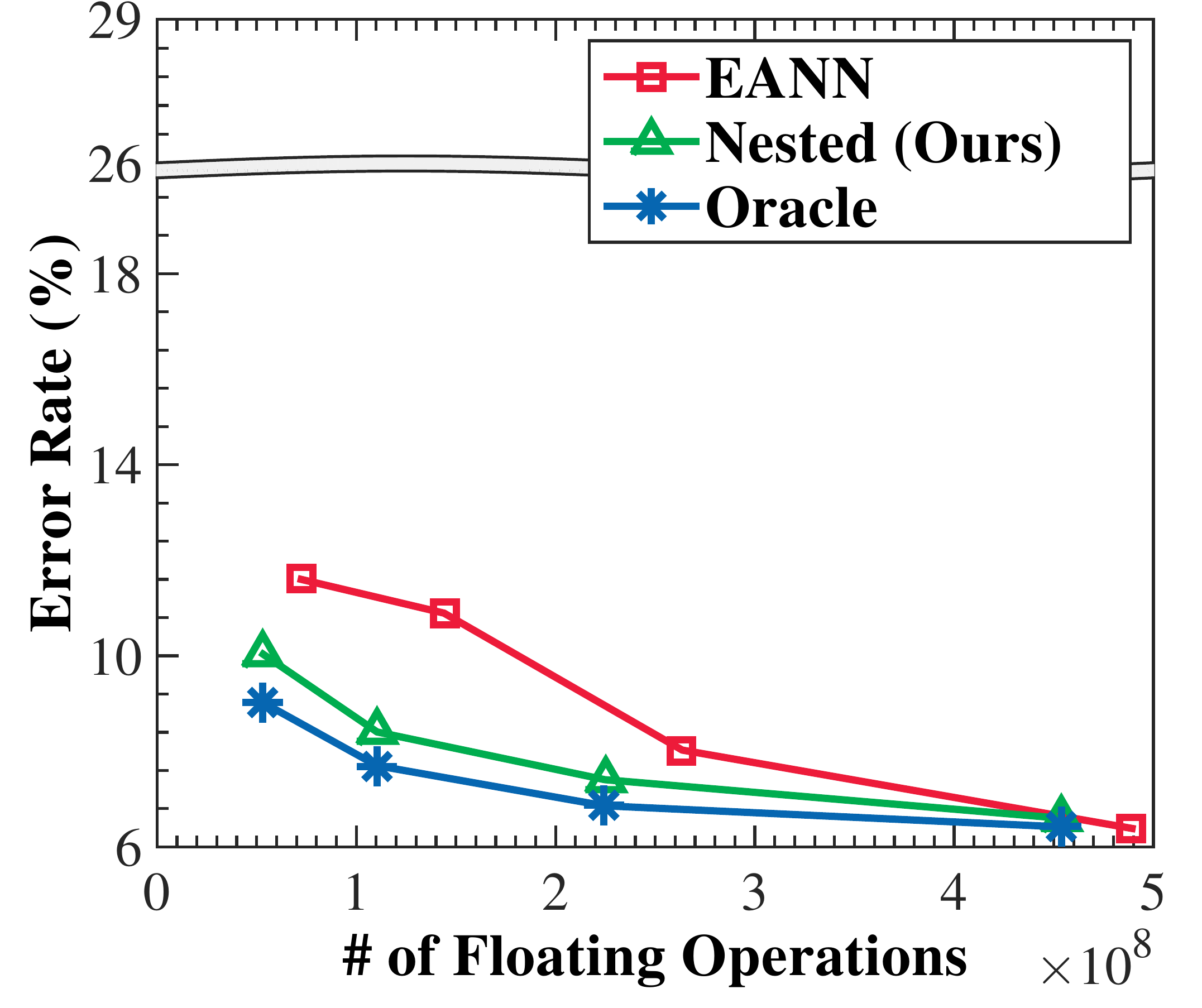}\\
         {\footnotesize{(a) Depth-nested;\\Sparse ResNet-98}}
      \end{minipage}
      \hfill
      \begin{minipage}[t]{0.32\linewidth}
         \vspace{0pt}
         \centering
         \includegraphics[width=1.0\linewidth]{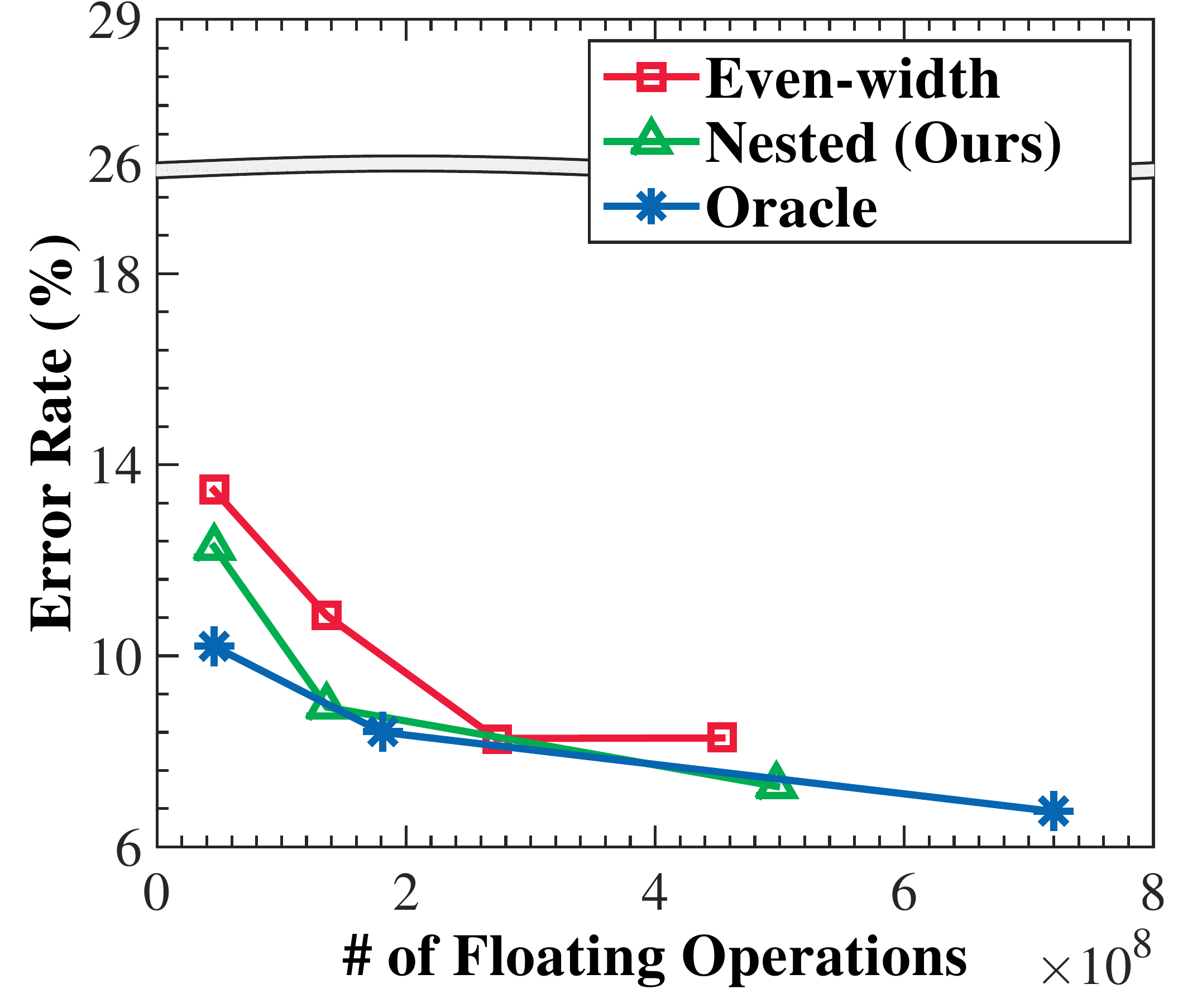}\\
         {\footnotesize{(b) Width-nested;\\ResNet-42}}
      \end{minipage}
      \hfill
      \begin{minipage}[t]{0.32\linewidth}
         \vspace{0pt}
         \centering
         \includegraphics[width=1.0\linewidth]{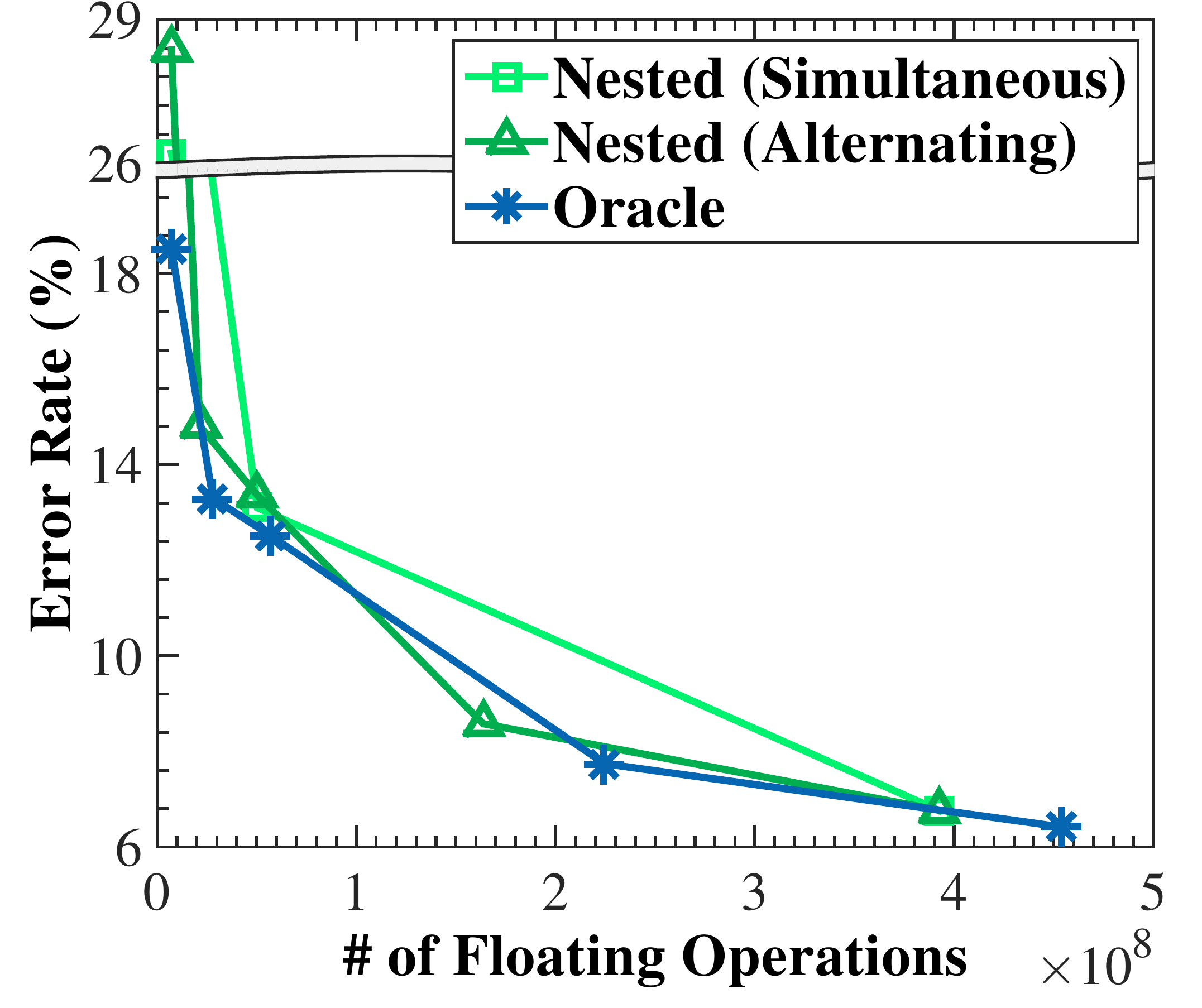}\\
         {\footnotesize{(c) Width-depth;\\Sparse ResNet-98}}
      \end{minipage}
      \vspace{-5pt}
      \caption{%
         Accuracy-FLOP trade-offs (lower is better).  Our nested architectures
         offer trade-offs close to the infeasible Oracle.%
      }%
      \label{fig:acc_time_curve_cnn}
   \end{minipage}
   \begin{minipage}{1.0\linewidth}
      \vspace{15pt}
      \centering
      \begin{minipage}[t]{0.32\linewidth}
         \vspace{0pt}
         \centering
         \includegraphics[width=1.0\linewidth]{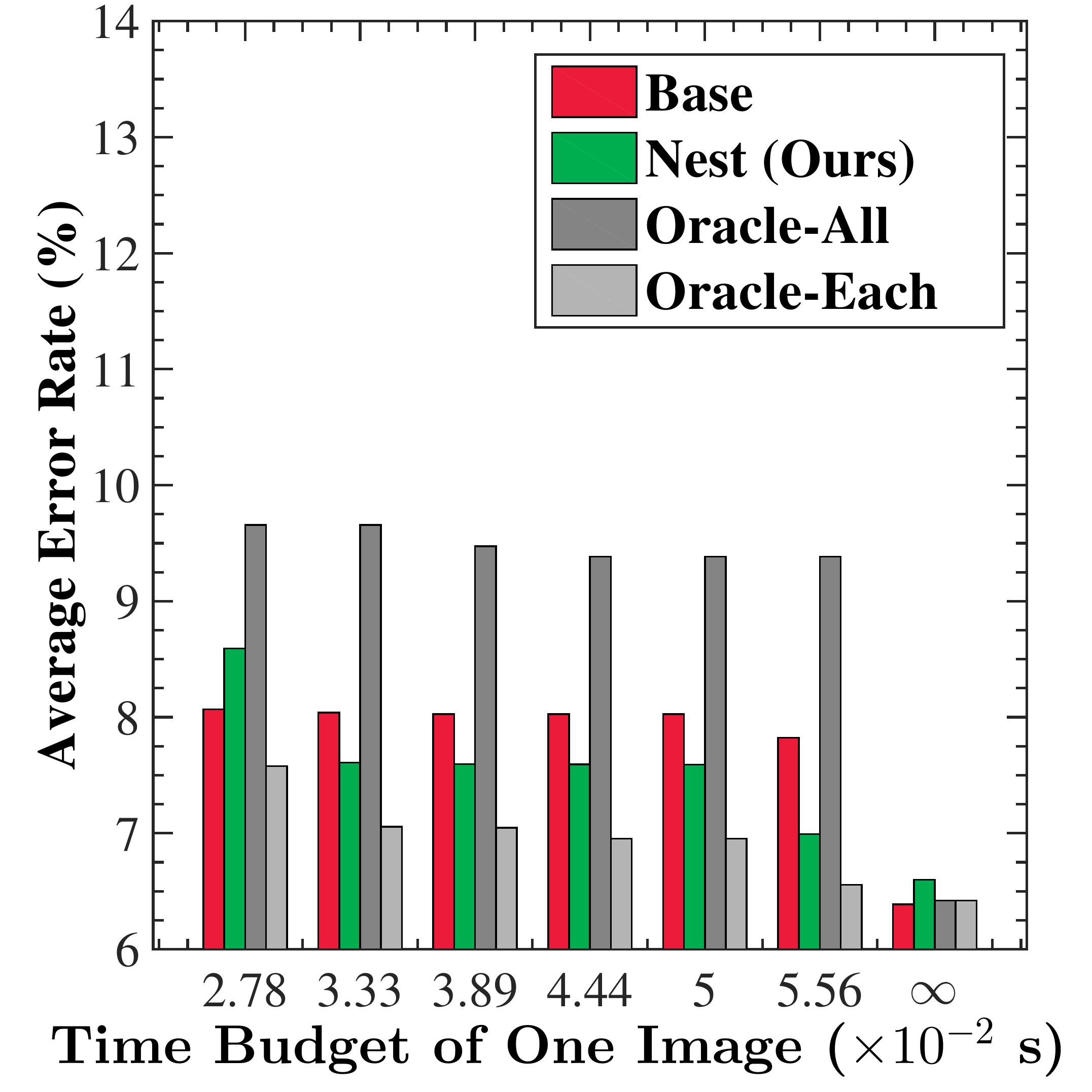}\\
         {\footnotesize{(a) Depth-nested;\\Sparse ResNet-98}}
      \end{minipage}
      \hfill
      \begin{minipage}[t]{0.32\linewidth}
         \vspace{0pt}
         \centering
         \includegraphics[width=1.0\linewidth]{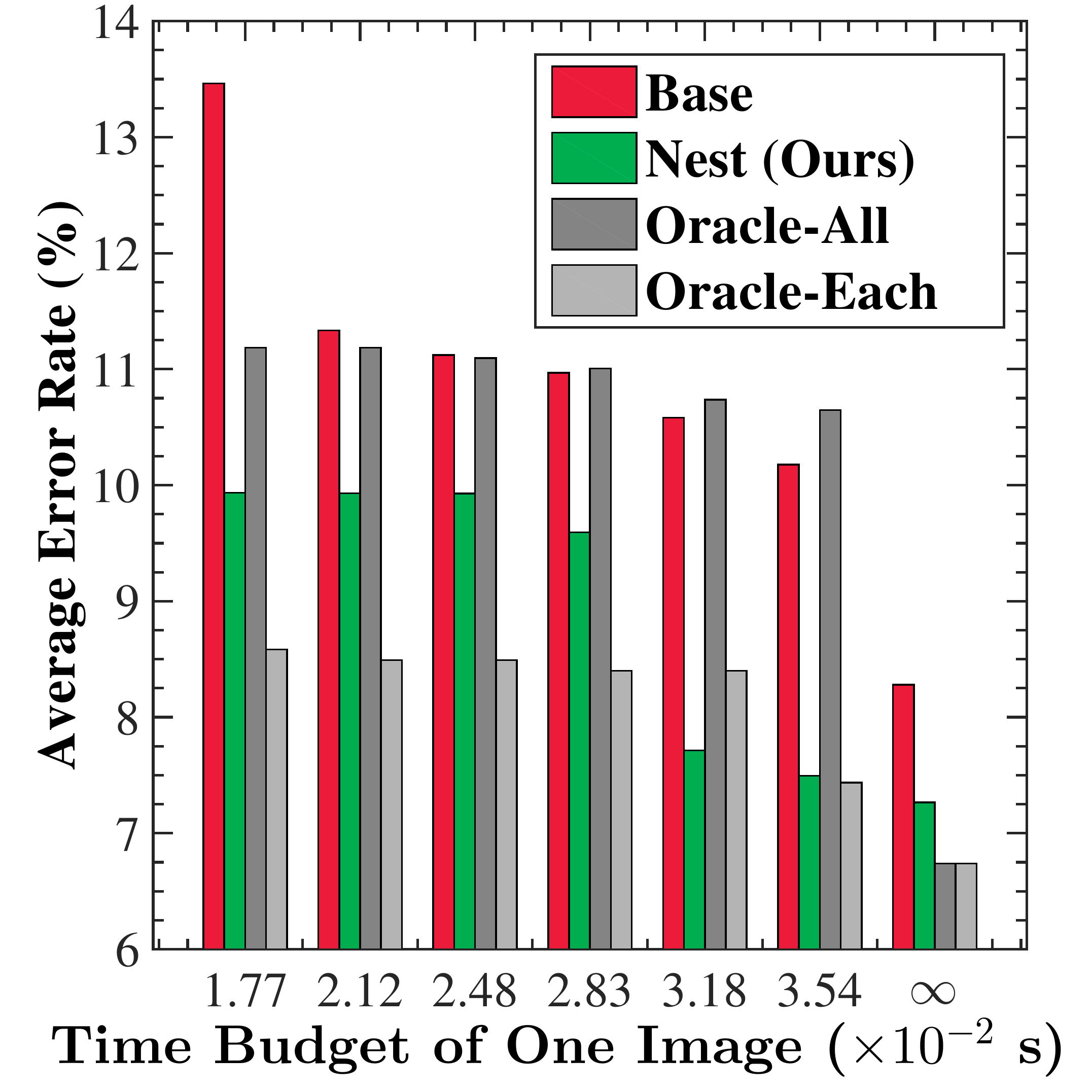}\\
         {\footnotesize{(b) Width-nested;\\ResNet-42}}
      \end{minipage}
      \hfill
      \begin{minipage}[t]{0.32\linewidth}
         \vspace{0pt}
         \centering
         \includegraphics[width=1.0\linewidth]{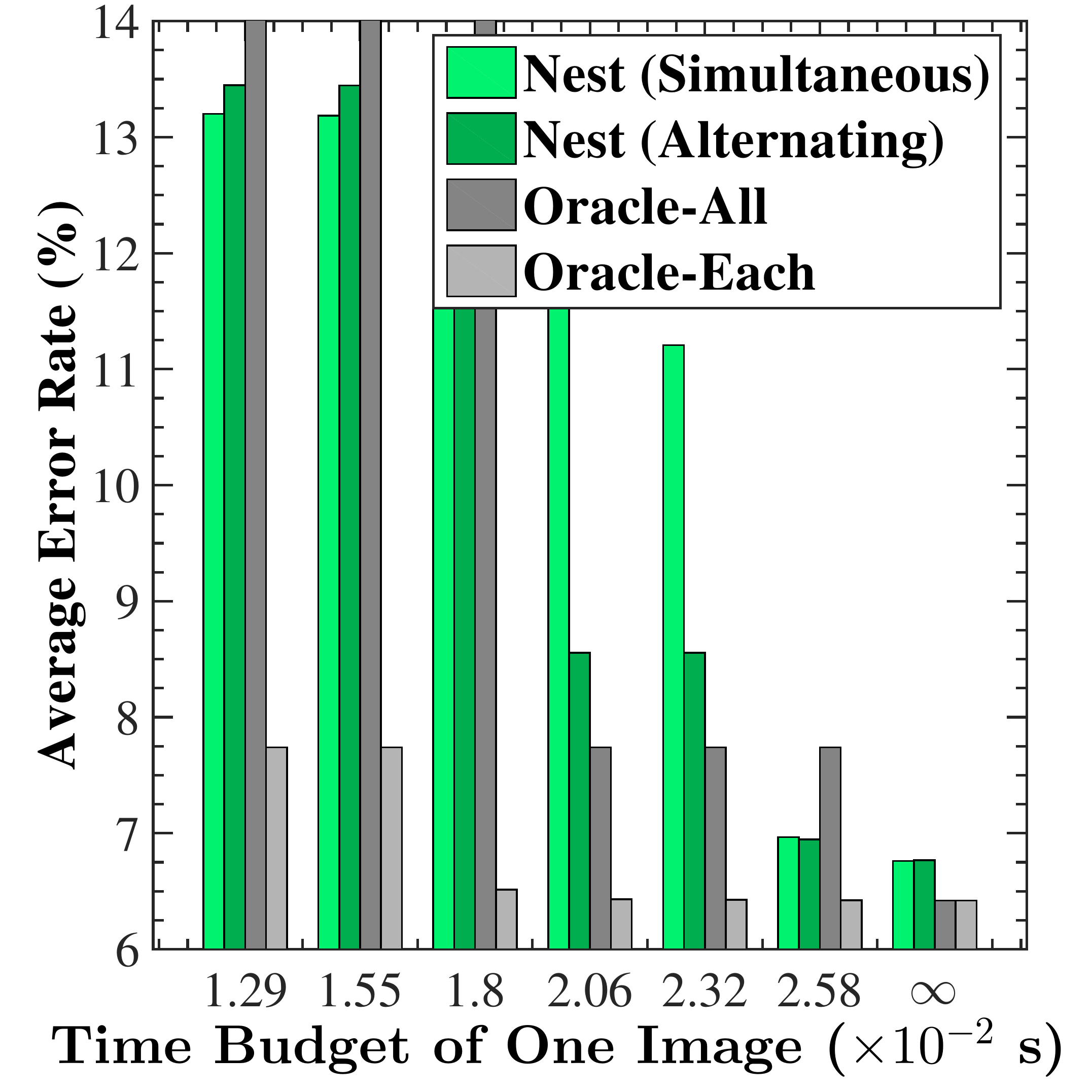}\\
         {\footnotesize{(c) Width-depth;\\Sparse ResNet-98}}
      \end{minipage}
      \vspace{-5pt}
      \caption{%
         Error rates at different deadlines (lower is better).  Our nested
         designs perform better than baselines and the static Oracle.%
      }
      \label{fig:acc_ddl_cnn}
   \end{minipage}
\end{figure}

We compare our nested architectures to an infeasible Oracle---a collection of
independently-trained single-task networks with sizes matching our subnetwork
stages.  Perfectly deploying this collection of independent networks as an
anytime system would require oracle knowledge of impending deadlines 
to select which network to run.  The Oracle thus represents an impossible
scenario in which anytime prediction capability is granted for free.
Figure~\ref{fig:acc_time_curve_cnn} shows the accuracy-FLOPs trade-off
curves achieved by our nested network designs (green), the Oracle (blue), and
the EANN and Even-width baselines (red).  Here, each network is trained using
the strategy that offers the most accurate results (\emph{i.e.}, OSGD for all
anytime networks and SGD for all independent networks except for the largest
setting of SparseResNet-98, which uses NormSGD).

From Figure~\ref{fig:acc_time_curve_cnn}a and~\ref{fig:acc_time_curve_cnn}b,
our depth and width nesting anytime networks both offer much better
accuracy-FLOPs trade-offs than previous work, and come close to the infeasible
Oracle.  Figure~\ref{fig:acc_time_curve_cnn}c shows our width-depth nested
Sparse ResNet-98 offers almost as good a trade-off as the Oracle, and covers
a much wider trade-off spectrum than depth-only or width-only nesting.

\subsection{Run-time Simulation}
\label{sec:exp_run_time}

We further compare four schemes for maximizing inference accuracy under various
inference deadlines:
(1) \textbf{Base}line anytime schemes (Even-width and EANN);
(2) Our \textbf{Nest}ed anytime schemes (width, depth, and width-depth nesting).
(3) \textbf{Oracle$_\text{All}$}, which picks the most accurate independent
network that finishes before the deadline for \textit{all} inputs;
(4) \textbf{Oracle$_\text{Each}$}, which picks the most accurate independent
network for each input that finishes before the deadline (\emph{i.e.}, the
network may vary across inputs).  When no inference result is generated by the
deadline, a random guess is output.  We report the average error rates across
all inputs in Figure~\ref{fig:acc_ddl_cnn} (vertical axis, lower is better)
under 7 deadlines and then no deadline (horizontal axis); the 7 deadlines are
set to be 0.5x-1x of the average latency under the biggest ResNet-42 or
Sparse ResNet across all inputs.

The accuracy advantage of \textbf{Nest} (the second bar in each group) over
\textbf{Base} (the first bar), and \textbf{Oracle}$_\text{All}$ (the third bar)
is apparent in Figure~\ref{fig:acc_ddl_cnn}.  For example, for ResNet-42,
\textbf{Nest} has 7\%-24\% lower error rate than \textbf{Base} for all
deadlines.  \textbf{Nest} has lower accuracy than \textbf{Oracle}$_\text{Each}$
in most cases, because the anytime network usually has slightly lower accuracy
than an independent network with same size.  Note that
\textbf{Oracle}$_\text{Each}$ is impractical, as it assumes impossible latency
prediction and no-overhead in swapping networks across inputs.  These
accuracy-under-deadline results are consistent with the accuracy-latency
curves in Figure~\ref{fig:acc_time_curve_cnn}.

\subsection{Evaluation on ImageNet}

Finally, we train a width-nested ResNet-50 and depth-nested Sparse ResNet-66
on the large-scale ImageNet (ILSVRC 2012) dataset \cite{imangenet}, using both
SGD and OSGD.  All networks are trained for 90 epochs, with learning rate
decreasing from 0.1 to 0.0001. Table~\ref{tab:training_imagenet} reports top-1
and top-5 validation error rates.  These results are consistent with our
previous findings on CIFAR.  On ImageNet, OSGD significantly improves the
accuracy of later stages (larger subnetworks) compared to standard SGD.

\begin{table}[t]
\centering
\small
\setlength{\tabcolsep}{5pt}
\begin{tabular}{c|ggcc}
\hline
      & \multicolumn{2}{g}{SGD}  & \multicolumn{2}{c}{OSGD} \\
      & Top-1 Error & Top-5 Error & Top-1 Error & Top-5 Error \\ \hline
\multicolumn{5}{c}{Our Width Nested ResNet-50}              \\ \hline
1$_{w1}$ & 36.7        & 14.7        & 36.7        & 14.8        \\
2$_{w2}$ & 31.5        & 11.7        & 31.7        & 11.7        \\
3$_{w4}$ & 29.2        & 10.2        & 28.3        & 9.4         \\ \hline
\multicolumn{5}{c}{Our Depth Nested Sparse ResNet-66}       \\ \hline
1$_{d1}$ & 31.3        & 11.3        & 32.9        & 12.4        \\
2$_{d2}$ & 28.4        & 9.7         & 29.2        & 10.1        \\
3$_{d4}$ & 28.0        & 9.3         & 27.1        & 8.9         \\ \hline
\end{tabular}
\vspace{-0.1in}
\caption{
   Validation error of anytime networks trained with SGD and OSGD on the
   ImageNet dataset.  As was the case for CIFAR-10
   (Table~\ref{tab:training_our}), OSGD improves the accuracy of later
   output stages.%
}%
\vspace{-0.2in}
\label{tab:training_imagenet}
\end{table}

\section{Conclusion}
\label{sec:conclusion}

We propose a new class of neural network architectures, which recursively
nest subnetworks in both width and depth.  We also propose \emph{Orthogonalized
SGD}, a novel variant of SGD customized for training such architectures with
re-balanced task-specific gradients.  We evaluate them with a variety of
network designs and achieve high accuracy and run-time flexibility.  Our
experiments demonstrate synergy between our architecture and optimizer: our
anytime networks perform almost as well as independent non-anytime networks of
the same size.

\textbf{Acknowledgments.}
This work is supported by NSF (grants CNS-1764039, CNS-1764039, CNS-1514256,
CNS-1823032), ARO (grant W911NF1920321), DOE (grant DESC0014195 0003), and the
CERES Center for Unstoppable Computing.  Continuing support for this line of
research is provided in part by NSF grant CNS-1956180.

\bibliography{citations}
\bibliographystyle{icml2020}

\end{document}